\documentclass[11pt,twoside]{article}
\usepackage[utf8]{inputenc} 
\usepackage[T1]{fontenc}    
\usepackage{hyperref}       
\usepackage{url}            
\usepackage{booktabs}       
\usepackage{amsfonts}       
\usepackage{nicefrac}       
\usepackage{microtype}      

\usepackage{float}
\usepackage[dvipsnames, svgnames, x11names]{xcolor} 
\usepackage{graphicx}
\usepackage{multirow}
\usepackage{subfigure}
\usepackage{geometry}
\usepackage{indentfirst}

\usepackage{amsmath}
\usepackage{amsthm}
\usepackage{amssymb}
\usepackage{color}

\usepackage{changepage}
\usepackage{caption}
\usepackage{xfrac}
\usepackage{txfonts}
\usepackage{enumitem}
\usepackage{diagbox} 

\usepackage{makecell}
\usepackage{textcomp}
\usepackage{array} 
\usepackage{longtable}
\usepackage{float}
\usepackage[justification=centering]{caption}

\usepackage{xcolor}

\usepackage{lineno}
\usepackage{comment}
\usepackage{setspace}

\begin{document}

\title{A Cross-Residual Learning for Image Recognition}

\author{Jun Liang \thanks{Corresponding Author. Email address: liangjun@m.scnu.edu.cn.}, Songsen Yu, and Huan Yang 
}

\date{}
\maketitle

\begin{abstract}
ResNets and its variants play an important role in various fields of image recognition. This paper gives another variant of ResNets, a kind of cross-residual learning networks called C-ResNets, which has less computation and parameters than ResNets. C-ResNets increases the information interaction between modules by densifying jumpers and enriches the role of jumpers. In addition, some meticulous designs on jumpers and channels counts can further reduce the resource consumption of C-ResNets and increase its classification performance. In order to test the effectiveness of C-ResNets, we use the same hyperparameter settings as fine-tuned ResNets in the experiments.

We test our C-ResNets on datasets MNIST, FashionMnist, CIFAR-10, CIFAR-100, CALTECH-101  and SVHN. Compared with fine-tuned  ResNets, C-ResNets not only maintains the classification performance, but also enormously reduces the amount of calculations and parameters which greatly save the utilization rate of GPUs and GPU memory resources. Therefore,  our C-ResNets is competitive and viable alternatives to ResNets in various scenarios. Code is available at  \url{https://github.com/liangjunhello/C-ResNet}

$\newline$\noindent\textbf{Key words}: C-ResNet, image recognition,  resources cost, jumper densification, dashed jumper
\end{abstract}

\section{Introduction}

Convolutional  (\emph{Conv}) neural networks  have made great achievements in computer vision in recent years.
For example, AlexNet \cite{krizhevsky2012imagenet} made a historic breakthrough in the field of conv neural networks in 2012. 
VGG \cite{simonyan2014very} used an architecture with $3\times3$ convolution filters to increase the network depth to 16\(\sim\)19 layers. 
EfficientNet \cite{tan2019efficientnet} provided a method for uniformly scaling all dimensions of depth/width/resolution.
In recent years, designing a compositional and adaptable framework  for image recognition has become a topic drawing the attention of researchers from various domains.

Broadening and  deepening the conv layer are two important strategies in the design of  conv neural networks. The Inception architecture of GoogLeNet \cite{szegedy2015going} expanded the width of the conv layer to enhance accuracy. Subsequent versions of Inception \cite{szegedy2016rethinking, szegedy2017inception} improved the performance of  network through a series of optimizations, such as adding batch normalization (\emph{BN}), turning the $7 \times7$ conv  kernel into a two-layer tandem of a $1 \times 7$ and a $7 \times1$ kernels  and introducing the idea of residual network into the model.
Residual network  (ResNet)~\cite{he2016deep} used a residual block layer and can gain accuracy through increasing the depth of conv layers,  without performance degradation or overfitting.
The building block of ResNeXt \cite{xie2017aggregated} broadens the residual block in ResNet, enriching the diversity of features.
SENet architectures~\cite{hu2018squeeze} based on Squeeze-and-Excitation (SE) block generalized effectively across different datasets. SE module increased channel's depth in ResNet and enhanced
channel-wise feature responses by  distinguishing  interdependencies between channels.
SKNets~\cite{li2019selective} were constructed through stacking multiple Selective Kernel (SK) units and using softmax attention to fuse multiple branches with different kernel sizes. SK unit expands the channels of the block of ResNeXt and SE, and SKNets can also gain the accuracy by deepening the network, that is, stacking more layers of SK units.  All the afore-mentioned variants of the original ResNet reveal that deepening and widening the conv layers are effective measures for performance enhancements. 

However, deepening and widening the conv layers is a big challenge for resource consumption. Therefore, we do not pay attention to these two aspects during the design of our residual networks. What we are concerned about is how to refine the design of ResNets so that its performance does not decrease while reducing its resource consumption. Our residual networks use the same width and less depth than ResNets mainly due to the design of cross jumpers.  Our experiments on six datasets show that our cross-residual networks (C-ResNets) have not only  fewer parameters and fewer computational costs, but also higher classification accuracy on most datasets than the corresponding fine-tuned  ResNets.

\section{Related Work}

In neural networks, the concept of residuals can also be understood as the reuse of historical information (memory). 
LSTM \cite{hochreiter1997long} or GRU \cite{cho2014learning} filtered historical experience through retaining some valuable memories and discarding some useless information.
To make better use of historical information, ResNet \cite{he2016deep} was proposed through simply stacking  residual blocks (or residual bottlenecks)  together to achieve better accuracy. Later ResNet was improved by changing the order of conv layer, BN and ReLu,  from conv layer $\Rightarrow$ BN $\Rightarrow$ ReLu   to BN $\Rightarrow$ ReLu $\Rightarrow$ conv layer  \cite{he2016identity}.

X-ResNet \cite{jou2016deep} used residual learning for multi-task cross-learning, which reduced the amount of parameters by more than 40\%.  X-ResNet achieved better detection performance on a visual sentiment concept detection problem normally solved by multiple specialized single-task networks.

ResNeXt \cite{xie2017aggregated} was a homogeneous, multi-branch architecture that had only a few hyper-parameters to set, which got 2nd place in the ILSVRC 2016 classification task. ResNeXt  repetitively employed building blocks that aggregated some transformations with the same size, which expanded the width of basicblock in \cite{he2016deep}.

In the building blocks of DenseNet \cite{huang2017densely}, each layer passed its own information to subsequent layers in a feed-forward fashion. DenseNet alleviated the vanishing-gradient problem, strengthened feature propagation, encouraged feature reuse, and substantially reduced parameter count.

The idea of cross-residual learning has also been applied in a very limited way in DCRNet \cite{shen2018deep}  for HEp-2 cell classification to enhance information transfer between different layers. DCRNet won the first place in the  ICPR 2016 contest.

ResNeSt  \cite{zhang2020resnest}  was formed by stacking Split-Attention blocks  in a ResNet-style manner, which outperformed other networks with similar model complexities. ResNeSt-50 achieved 81.13\% top-1 accuracy on ImageNet  only using a single crop-size of $224 \times 224$.

ConvNeXt \cite{liu2022convnet} were dubbed and constructed entirely by standard residual conv modules. It was competitive compared with Transformer \cite{dosovitskiy2020image, he2022masked} in terms of accuracy and scalability and  achieved 87.8\% top-1 accuracy on ImageNet.

A residual multiscale module with an attention mechanism (RMAM) \cite{lan2020madnet}  effectively extracted multiscale features and utilized the discriminative information among different channels. A dual residual-path block (DRPB)  \cite{lan2020madnet} adopted the hierarchical features from low-resolution images which was comprised of a dense lightweight network (MADNet) for stronger multiscale feature expression and feature correlation learning.

A cascading residual network (CRN) \cite{lan2020cascading} and an enhanced residual network (ERN) \cite{lan2020cascading} were proposed to separately  promote the propagation of feature and enhance image resolution. And we can also know that the combination of multiscale blocks and residual networks is effective in the single-image super-resolution (SISR) field.

In \cite{chen2018embedding},  a residual refinement network introducing a second-order term into element-wise addition was applied to fuse the learned multilevel features for salient object detection.

We can see that ResNets and its variants can perform better in various areas of computer vision. Residual networks can achieve better performance by fully exploiting residual information.
However, the stacking pattern of modules in most variants of ResNets is different from ResNets, or some variants just use the idea of residual module. This paper intends to redesign the residual module in the residual network and stacks it in the ResNet-style so that  our residual networks can maximize the use of residual information and maintain the diversity of information (i.e., the diversity of features) in different locations to reduce the cost of network.

\section {The Proposed Cross-Residual Network Architecture}

ResNet (v2) \cite{he2016identity} discussed experimentally the effects of Addition before and after the ReLu function and showed that  placing Addition before the ReLu function is better than putting it after ReLu.  However, we wonder if a more elegant framework exists which regards conv layer, BN, and ReLu as a whole, despite conflicting with the experimental result, and if the  framework can be applied to the residual network. Suppose we treat conv layer, BN and ReLu as a whole. How should we design the order of conv layer, BN and ReLu?  Moreover, our ReLu function will be designed before the Addition which is contrary to the experimental result of \cite{he2016identity}. Is this conjecture valid? We shall discuss it in Section (\ref{Experiments}). Below we first present the WBR layer consisting of a sequence of Weight layer (i.e., conv layer), Batch normalization, and ReLu and cross building blocks.  Then we show the different structures of C-ResNets and their implementation details.

\begin{figure*}[htb]
 \subfigure[~]{
  \label{wbr-A}      
  \begin{minipage}[b]{0.15\textwidth}
  	\centering
 	\includegraphics[width=0.87in]{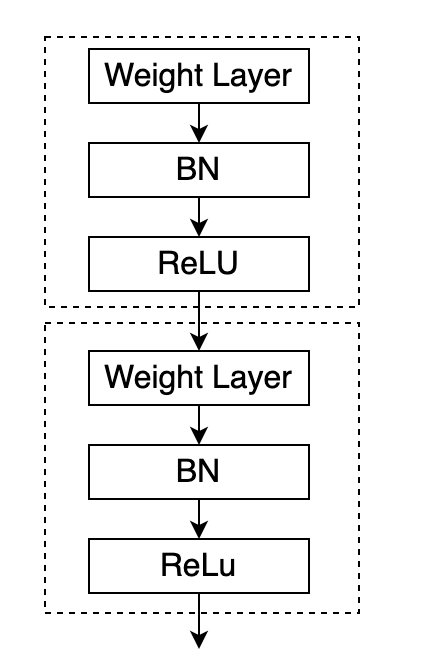}
  \end{minipage}}%
   \subfigure[~]{
  \label{wbr-A}      
  \begin{minipage}[b]{0.15\textwidth}
  	\centering
 	\includegraphics[width=0.87in]{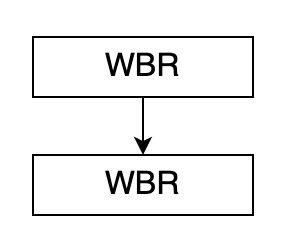}
  \end{minipage}}%
 \subfigure[~]{
  \label{wbr-A}      
  \begin{minipage}[b]{0.15\textwidth}
  	\centering
 	\includegraphics[width=0.87in]{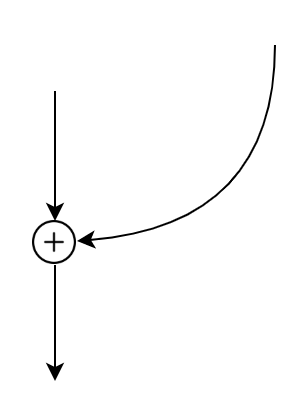}
  \end{minipage}}%
\subfigure[~]{
  \label{wbr-B}       
  \begin{minipage}[b]{0.15\textwidth}
   \centering
   \includegraphics[width=0.85in]{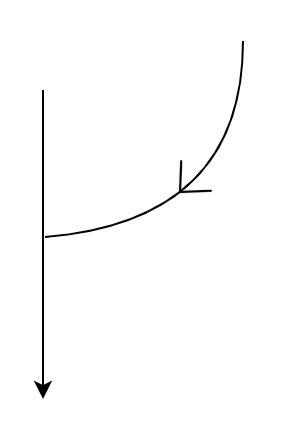}
  \end{minipage}}
   \subfigure[~]{
  \label{three arrow-1}      
  \begin{minipage}[b]{0.15\textwidth}
   \centering
   \includegraphics[width=0.8in]{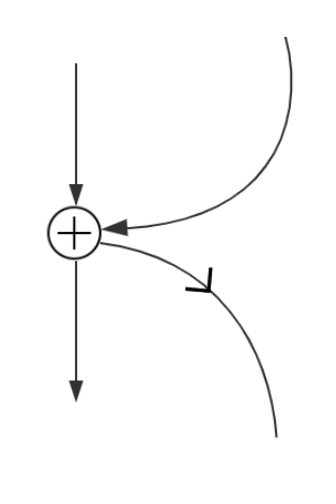}
  \end{minipage}}%
  \subfigure[~]{
  \label{three arrow-2}       
  \begin{minipage}[b]{0.15\textwidth}
   \centering
   \includegraphics[width=0.8in]{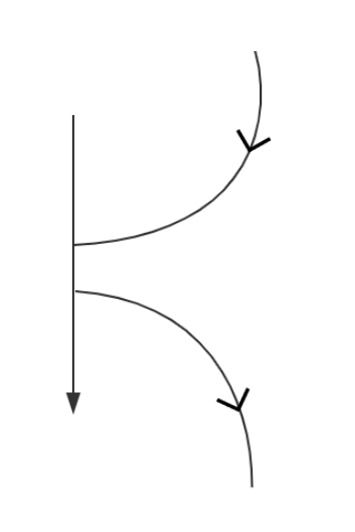}
  \end{minipage}}
 \caption{ (b) is WBR representation of (a). (d) and (f) are simplified representations of (c) and (e) without  addition sign, respectively.}
 \label{Wbr} 
\end{figure*}

\subsection{The cross building blocks and their WBR Layer representations}

\textbf{WBR layer.}
A WBR layer is designed to contain a  series of operations following the order of conv, BN and then ReLu.  In Fig. (\ref{Wbr}), (b) is the WBR representation of (a); (d) and (f) are simplified notations of (c) and (e), respectively.

Let the output of a WBR layer with input \(x\) be $P (x)$. 
Hence, we have
\begin{equation} 
P(x)=\sigma(B(W(x))),
\end{equation}
where W, B and $\sigma$ separately denote Weight layer, Batch normalization layer and ReLu function.

\begin{figure*}[!h]
	\centering
	\begin{adjustwidth}{-0in}{0in}
		\centering
		\subfigure[~]{
		\label{Original-Bottleneck-D}
		\begin{minipage}[t]{0.25\linewidth}
			\centering
			\includegraphics[width=0.82in]{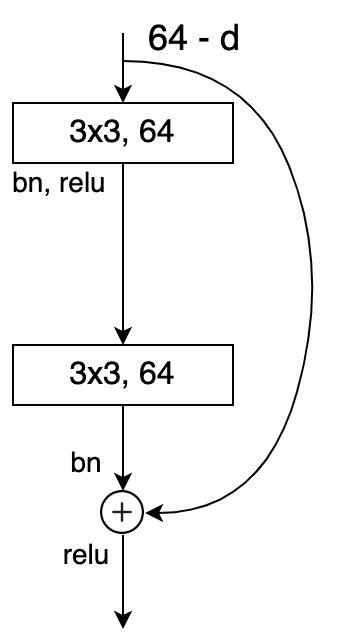}
		\end{minipage}%
		}%
		\subfigure[~]{
		\label{Cross-Block-A}
		\begin{minipage}[t]{0.25\linewidth}
			\centering
			\includegraphics[width=0.86in]{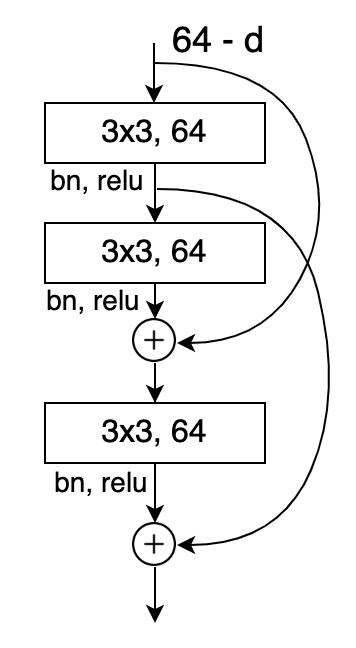}
			\end{minipage}%
		}%
		 \subfigure[~]{
 		 \label{wbr-A}      
 		 \begin{minipage}[b]{0.3\textwidth}
   			 \centering
 		 	\includegraphics[width=1.2in]{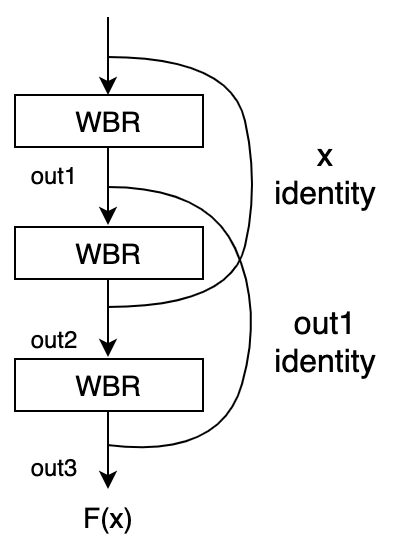}
 		 \end{minipage}}%
		
		\subfigure[~]{
		\label{Original-Bottleneck-E}
		\begin{minipage}[t]{0.2\linewidth}
			\centering
			\includegraphics[width=0.93in]{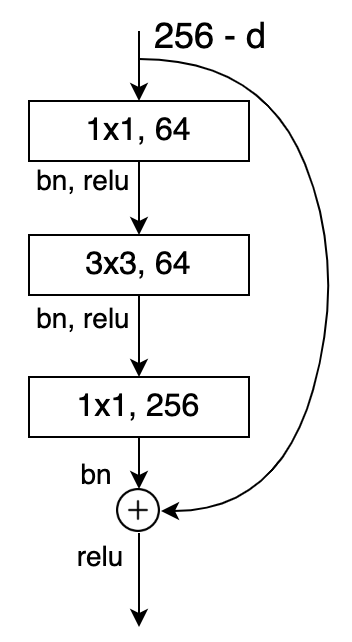}
		\end{minipage}%
		}%
		\subfigure[~]{
		\label{Cross-Bottleneck-C}
		\begin{minipage}[t]{0.2\linewidth}
			\centering
			\includegraphics[width=0.88in]{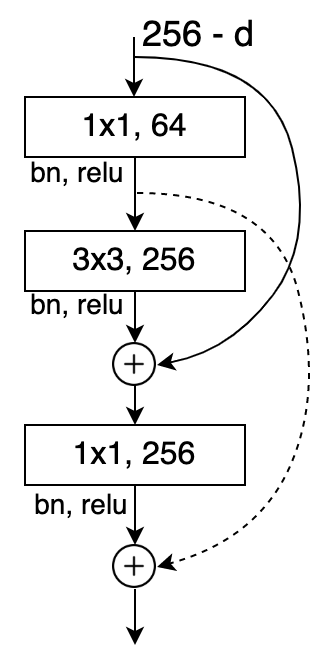}
		\end{minipage}%
		}%
		\subfigure[~]{
		\label{Cross-Bottleneck-B}
		\begin{minipage}[t]{0.2\linewidth}
			\centering
			\includegraphics[width=0.695in]{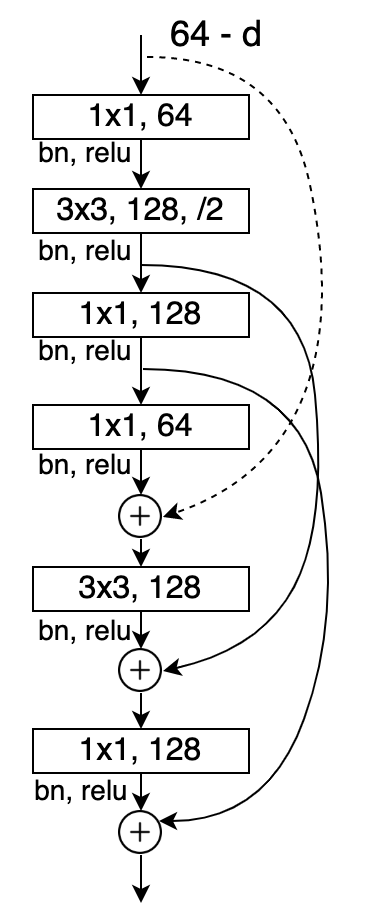}
		\end{minipage}%
		}%
		\subfigure[~]{
  		\label{wbr-B}       
  		\begin{minipage}[b]{0.28\textwidth}
   			\centering
   			\includegraphics[width=0.91in]{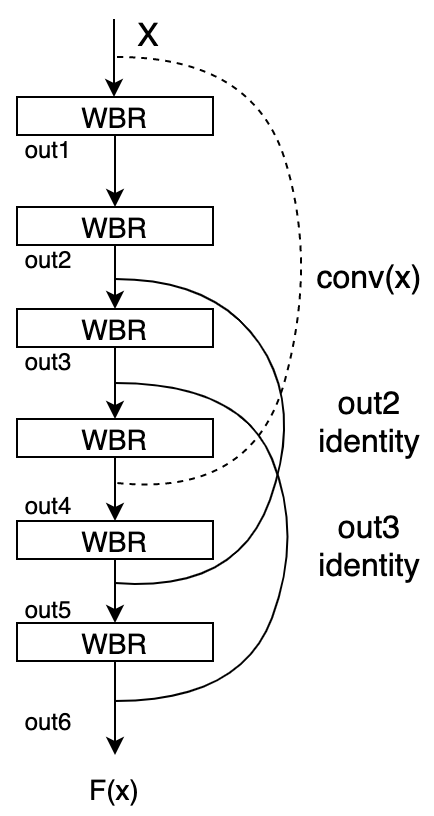}
  		\end{minipage}%
  		}%
	\end{adjustwidth}
	\begin{center}
		\caption{ (a).Original BasicBlock in ResNet  \ (b). Our Cross-Block  \ (c). WBR representation of (b) \ (d). Original Bottleneck in ResNet  \ (e). Our three-layer Cross-Bottleneck \ (f). Our  six-layer Cross-Bottleneck \  (g). WBR representation of (f).}
		\label{basic and improved block}
	\end{center}
\end{figure*}

\textbf{Cross-Block.} A Cross-Block (Fig. \ref{Cross-Block-A})  has three WBR layers and two cross skip connection. The purpose of crossing the two jumpers is to utilize more jumpers in fewer WBR layers so that  the transfer of information among layers is enhanced.  When the skip connection is a dashed jumper, it means that a $ 1 \times1 $ conv operation needs to be done. Here we give two variants for Cross-Block, which are  the cases that the first skip connection is a dashed jumper and that the second skip connection is a dashed jumper, separately called Cross-Block-A2 and Cross-Block-A1. To differentiate these variants, Fig. \ref{Cross-Block-A} is also called Cross-Block-A. The WBR representation (Fig. \ref{wbr-A}) of Cross-Block-A can be expressed as:
\begin{equation} \label{formu_improve_block}
\begin{split}
out1=P(x),
 out2=P(out1)+x,\\
 out3=P(out2)+out1,\\
F(x)=out3=P(P(P(x))+x)+P(x).
\end{split}
\end{equation}

The notation ``+''  means an element-wise addition operation. Like the ResNet model, the ``+'' operation does not produce additional parameters or computation cost. 

Generally, let $P^m(x)$ and $P_c^m(x)$ roughly represent the nesting of $m \  (m \ge 2)$ $p$-functions. For example, $P^2(x)=P(P(x)+x)$ and $P_c^2(x)=P(P(x)+conv(x))$. Then the output of  Cross-Block-A can be rewritten as $P^3(x)+P(x)$.
Similarly, the outputs of Cross-Block-A1 and Cross-Block-A2 are $P_c^3(x)+P(x)$ and  $P^3(x)+conv(P(x))$, respectively.

\textbf{Cross-Bottleneck.}  A Cross-Bottleneck has three layers and two skip connections (Fig. \ref{Cross-Bottleneck-C}) or six layers and three skip connections (Fig. \ref{Cross-Bottleneck-B}). A six-layer Cross-Bottleneck  is a stack of two three-layer  Cross-Bottlenecks, which is  designed in order to keep the number of skip connections (solid) as much as possible while reducing the number of channels (from 256 to 128) facilitating a significant reduction in the amount of calculation. Applying a similar strategy, the WBR representation (Fig. \ref{wbr-B}) of six-layer Cross-Bottleneck can be expressed as: $F(x)= P(P_c^5(x)+P^2(x))+P^3(x).$

  \begin{figure*}[htb]
   \subfigure[ResNet18 and fine-tuned ResNet18]{
  \label{ResNet18}      
  \begin{minipage}[b]{0.22\textwidth}
   \centering
   \includegraphics[width=1.515in]{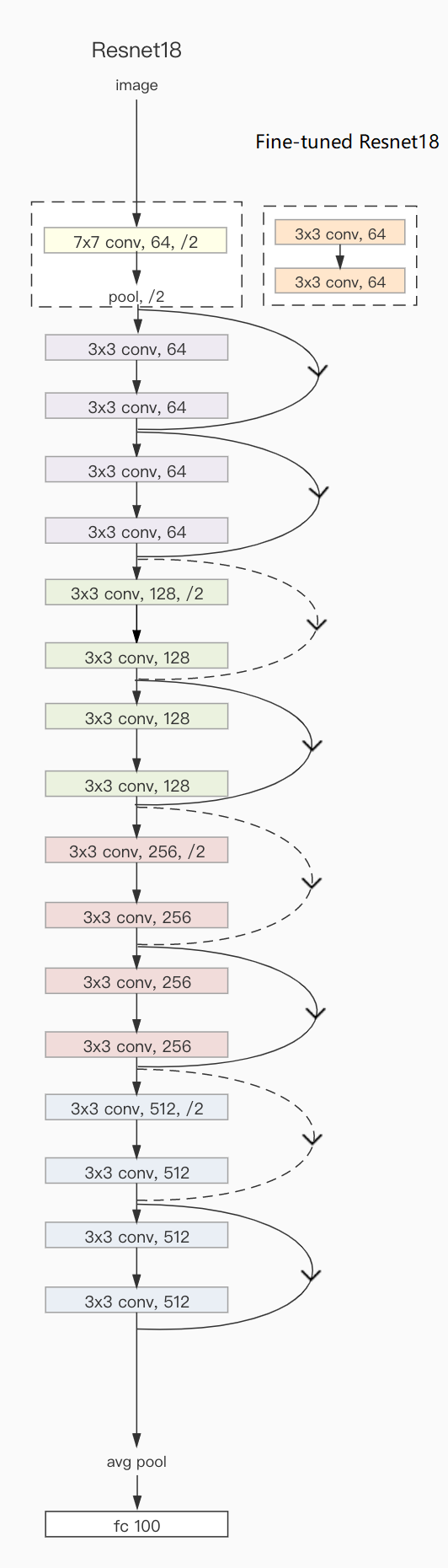}
  \end{minipage}}%
   \subfigure[C-ResNet18-A (or, C-ResNet18)]{
  \label{C-ResNet18-A}      
  \begin{minipage}[b]{0.18\textwidth}
   \centering
   \includegraphics[width=1.45in]{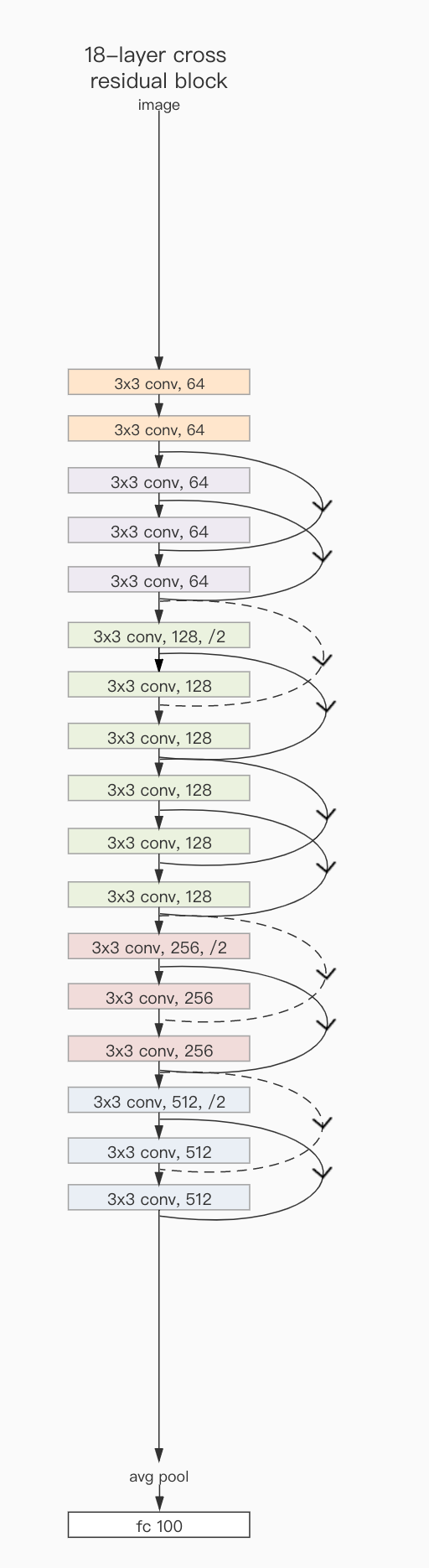}
  \end{minipage}}%
      \subfigure[C-ResNet27-A]{
  \label{C-ResNet27-A]}      
  \begin{minipage}[b]{0.18\textwidth}
   \centering
   \includegraphics[width=1.45in]{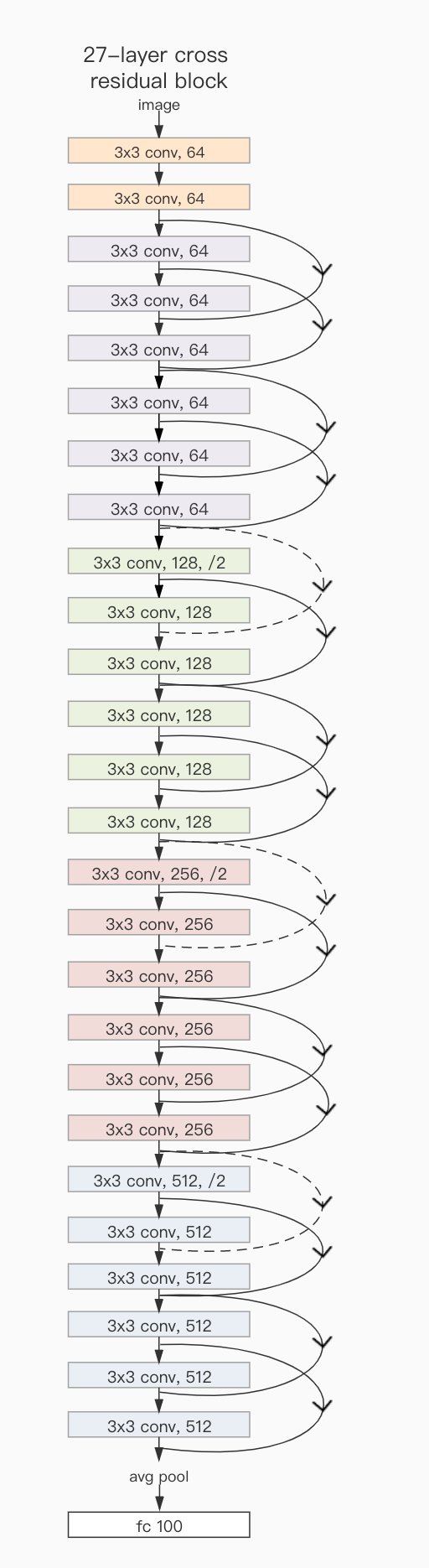}
  \end{minipage}}%
\subfigure[C-ResNet27-A1]{
  \label{C-ResNet27-A1}       
  \begin{minipage}[b]{0.18\textwidth}
   \centering
   \includegraphics[width=1.45in]{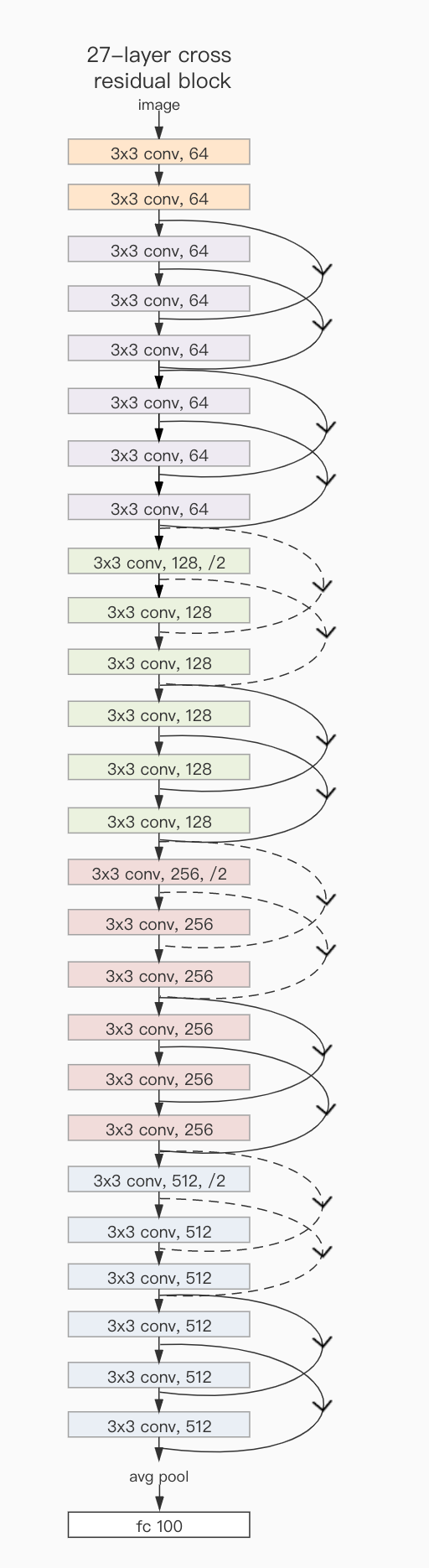}  
  \end{minipage}}%
    \subfigure[C-ResNet27-A2]{
  \label{C-ResNet27-A2}       
  \begin{minipage}[b]{0.18\textwidth}
   \centering
   \includegraphics[width=1.45in]{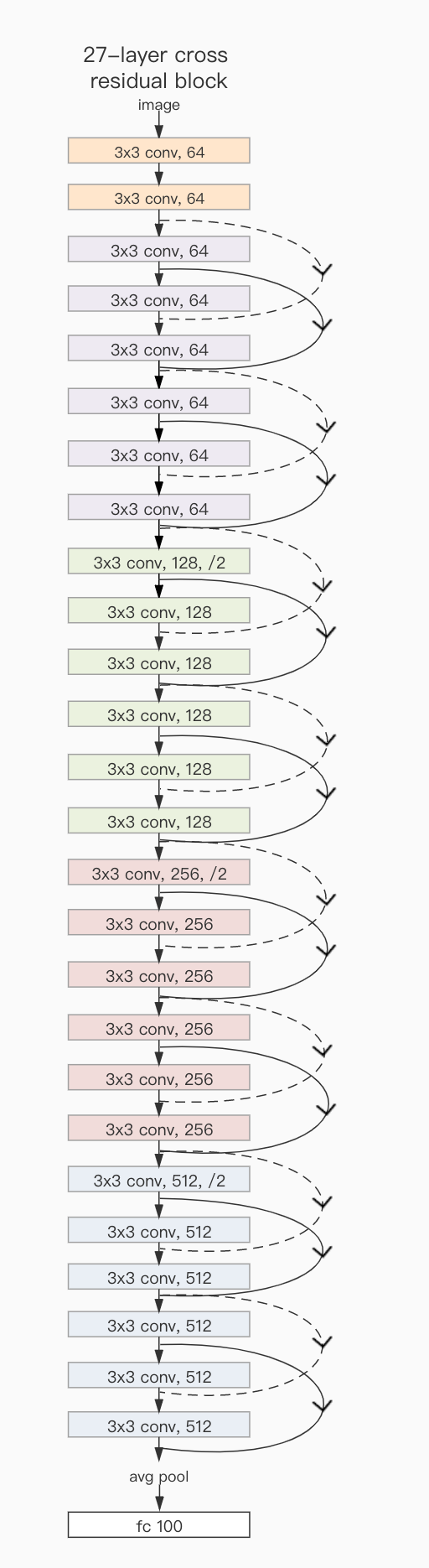}      
    \end{minipage}}%
 \caption{Fine-tuned ResNet18 (actually 19 layers) and samples for C-ResNets.  According to the position of the dashed skip connections, three variants (c), (d) and (e) of C-ResNet27 are constructed. } 
 \label{network architecture-1} 
\end{figure*}

\textbf{Discussion.} Two types of cross-residual learning modules are proposed which can be stacked to form cross-residual learning networks (i.e., C-ResNets). The differences between our cross-residual modules and residual modules in ResNets \cite{he2016deep, he2016identity} are: (a) the location of Addition (our Addition is after ReLu, not before as modules in  ResNets); (b) the number of convolutional layers and skip connections (compared  with original blocks,  we insert additional the convolutional layers in order to increase the skip connection number and make them interweave); (c)  the role of  $1 \times1$ convolution operation on skip connection (when we use skip connection,  a $1 \times1$ convolution operation may be done to increase the transformation of features, regardless of whether the dimensions at both ends of the skip connection match); (d) the channel counts (compared with original Bottleneck, our Cross-Bottleneck reduces the channel counts while maintaining classification performance).

\subsection{Network Architectures}
Cross residual networks can be constructed by separately stacking these cross residual modules.  These networks can be named according to its residual module and its number of layers. For example, C-ResNet18-A (Fig.  \ref{C-ResNet18-A}) is stacked by Cross-Block-A and  its layers number of 18.  When there is no ambiguity, we also write C-ResNet18-A as C-ResNet18, and C-ResNet15-A1 as C-ResNet15. In order to enable ResNet (applied on ImageNet dataset and only processing $224 \times 224$ pictures) to process $32 \times 32$ pictures, we have fine-tuned ResNet (called ResNet*)  (Fig.  \ref{ResNet18}), where the $7 \times7$ conv and maximum pooling are  replaced with two $3 \times 3$ conv layers, and different numbers of neurons in the fully connected layer  are set to adapt to the number of categories in different datasets. In Section \ref{Experiments}, we show that the classification accuracy of  ResNet* is higher than that of  ResNet (v2)  \footnote{The code \cite{he2016identity} is available at  \url{https://github.com/facebookarchive/fb.resnet.torch/blob/master/models/preresnet.lua}} on datasets CIFAR-10 and CIFAR-100. Hence, we use fine-tuned ResNet (i.e., ResNet*) as the baseline for image recognition task.  For different tasks,  we use different items for our C-ResNets.  The input of object detection task is still an image of size  $224 \times 224$ , the item of C-ResNet still employ the $7 \times7$ conv and maximum pooling. In addition, the execution settings and peripheral environment of ResNet*  (or ResNet) and C-ResNet is exactly the same except for the stacked modules such that the fairness of our comparative experiments is guaranteed.

 C-ResNet-B and C-ResNet-C are separately stacked by  six-layer Cross-Bottlenecks and three-layer Cross-Bottlenecks. 
Table  \ref{The architectures for our C-ResNet and ResNet*}  shows the specific information on the structures of C-ResNets and  fine-tuned ResNets including numbers of dashed skip connections and their corresponding FLOPs. Our C-ResNet15 (or C-ResNet18), C-ResNet27-A2 and C-ResNet27-B2 are designed to compare with  fine-tuned ResNet18, ResNet34 and ResNet50, respectively. It can be seen that the FLOPs of C-ResNets are fewer than the corresponding ResNets. Why do we compare like this? For example, why do we compare the 27-layer C-ResNet-A2 with the 34-layer fine-tuned ResNet (i.e., ResNet34*)? Although C-ResNet27-A2 has fewer layers than ResNet34*, it has five more dashed skip connections than ResNet34*, which means that  five more $1\times1$ conv operations are done. Hence, compared with ResNet34*, C-ResNet27-A2 can be regarded as using skip connections in exchange for conv layers (equivalent to 32 conv layers).

\setlength{\tabcolsep}{2mm}{ 
	\begin{table*}[!htbp]   \tiny 
		\caption{The architectures for our C-ResNet and ResNet*.  FLOPs include computational expenses  required by $1 \times 1$ conv on skip connections.   In addition, Con2\_1, Con3\_1, Con4\_1 and Con5\_1 take downsampling. The calculation of FLOPs and Params are based on the image of $32 \times 32$ and the dataset CIFAR-100.}
		\label{The architectures for our C-ResNet and ResNet*}
		\centering
		
		\begin{tabular}{|c|c|c|c|c|c|c|c|c|c|c}   
			\hline
			\multirow{2}*{layer name} & \multirow{2}*{output size}& Fine-tuned  &Fine-tuned  & Fine-tuned  &  C-ResNet15-A1&C-ResNet18-A & C-ResNet27-A2  &C-ResNet27-C1
			& C-ResNet27-B2  \\
			& &ResNet18 & ResNet34 &ResNet50 &  [1,1,1,1] & [1,2,1,1]& [2,2,2,2] &  [2,2,2,2] & [1,1,1,1]   \\
			\hline
			conv1 & 32×32 	&\multicolumn{8}{c|}{3×3, 64,stride 1}\\
			\hline
			conv2 & 32×32 	&\multicolumn{8}{c|}{3×3, 64,stride 1}\\

			\hline
		\multirow{6}*{conv2x} &
	    \multirow{6}*{32×32} &
		\multicolumn{1}{c|}{\multirow{6}{*}{	
				$
				\begin{bmatrix}
				3$×$3,64\\
				3$×$3,64\\
				\end{bmatrix}$×$2
				$}}&
		\multicolumn{1}{c|}{\multirow{6}{*}{
				$
				\begin{bmatrix}
				3$×$3,64\\
				3$×$3,64\\
				\end{bmatrix}$×$3
				$}}&
		
		\multicolumn{1}{c|}{\multirow{6}{*}{
				$
				\begin{bmatrix}
				1$×$1,64\\
				3$×$3,64\\
				1$×$1,256\\
				\end{bmatrix}$×$3
				$}}&
		\multicolumn{1}{c|}{\multirow{6}{*}{
				$
				\begin{bmatrix}
				3$×$3,64\\
				3$×$3,64\\
				3$×$3,64\\
				\end{bmatrix}$×$1
				$}}&
		\multicolumn{1}{c|}{\multirow{6}{*}{
				$
				\begin{bmatrix}
				3$×$3,64\\
				3$×$3,64\\
				3$×$3,64\\
				\end{bmatrix}$×$1
				$}}&
		\multicolumn{1}{c|}{\multirow{6}{*}{
				$
				\begin{bmatrix}
				3$×$3,64\\
				3$×$3,64\\
				3$×$3,64\\
				\end{bmatrix}$×$2
				$}}&
		\multicolumn{1}{c|}{\multirow{6}{*}{
				$
				\begin{bmatrix}
				1$×$1,64\\
				3$×$3,64\\
				1$×$1,256\\
				\end{bmatrix}$×$2
				$}}&
		\multicolumn{1}{c|}{\multirow{6}{*}{
					$
				\begin{bmatrix}
				1$×$1,64\\
				3$×$3,128\\
				1$×$1,128\\
				1$×$1,64\\
				3$×$3,128\\
				1$×$1,128\\
				\end{bmatrix}$×$1
				$}}
	\\
	~&~	~&~	~&~	~&~	~&~	~&~	~&~	~&~	~&~
	\\
	~&~	~&~	~&~	~&~	~&~	~&~	~&~	~&~	~&~
	\\ 
	~&~	~&~	~&~	~&~	~&~	~&~	~&~	~&~	~&~
	\\
	~&~	~&~	~&~	~&~	~&~	~&~	~&~	~&~	~&~
	\\ 
	~&~	~&~	~&~	~&~	~&~	~&~	~&~	~&~	~&~
	\\
	\hline
	\multirow{6}*{conv3x} &
	\multirow{6}*{16×16} &
	\multicolumn{1}{c|}{\multirow{6}{*}{	
			$
			\begin{bmatrix}
			3$×$3,128\\
			3$×$3,128\\
			\end{bmatrix}$×$2
			$}}&
	\multicolumn{1}{c|}{\multirow{6}{*}{
			$
			\begin{bmatrix}
			3$×$3,128\\
			3$×$3,128\\
			\end{bmatrix}$×$4
			$}}&
	
	\multicolumn{1}{c|}{\multirow{6}{*}{
			$
			\begin{bmatrix}
			1$×$1,128\\
			3$×$3,128\\
			1$×$1,512\\
			\end{bmatrix}$×$4
			$}}&
	\multicolumn{1}{c|}{\multirow{6}{*}{
			$
			\begin{bmatrix}
			3$×$3,128\\
			3$×$3,128\\
			3$×$3,128\\
			\end{bmatrix}$×$1
			$}}&
	\multicolumn{1}{c|}{\multirow{6}{*}{
			$
			\begin{bmatrix}
			3$×$3,128\\
			3$×$3,128\\
			3$×$3,128\\
			\end{bmatrix}$×$2
			$}}&
	\multicolumn{1}{c|}{\multirow{6}{*}{
			$
			\begin{bmatrix}
			3$×$3,128\\
			3$×$3,128\\
			3$×$3,128\\
			\end{bmatrix}$×$2
			$}}&
	\multicolumn{1}{c|}{\multirow{6}{*}{
			$
			\begin{bmatrix}
			1$×$1,128\\
			3$×$3,128\\
			1$×$1,512\\
			\end{bmatrix}$×$2
			$}}&
	\multicolumn{1}{c|}{\multirow{6}{*}{
			$
			\begin{bmatrix}
			1$×$1,128\\
			3$×$3,256\\
			1$×$1,256\\
			1$×$1,128\\
			3$×$3,256\\
			1$×$1,256\\
			\end{bmatrix}$×$1
			$}}
	\\
	~&~	~&~	~&~	~&~	~&~	~&~	~&~	~&~	~&~
	\\
	~&~	~&~	~&~	~&~	~&~	~&~	~&~	~&~	~&~
	\\ 
	~&~	~&~	~&~	~&~	~&~	~&~	~&~	~&~	~&~
	\\
	~&~	~&~	~&~	~&~	~&~	~&~	~&~	~&~	~&~
	\\ 
	~&~	~&~	~&~	~&~	~&~	~&~	~&~	~&~	~&~
	\\
	\hline
	\multirow{6}*{conv4x} &
	\multirow{6}*{8×8} &
	\multicolumn{1}{c|}{\multirow{6}{*}{	
			$
			\begin{bmatrix}
			3$×$3,256\\
			3$×$3,256\\
			\end{bmatrix}$×$2
			$}}&
	\multicolumn{1}{c|}{\multirow{6}{*}{
			$
			\begin{bmatrix}
			3$×$3,256\\
			3$×$3,256\\
			\end{bmatrix}$×$6
			$}}&
	
	\multicolumn{1}{c|}{\multirow{6}{*}{
			$
			\begin{bmatrix}
			1$×$1,256\\
			3$×$3,256\\
			1$×$1,1024\\
			\end{bmatrix}$×$6
			$}}&
	\multicolumn{1}{c|}{\multirow{6}{*}{
			$
			\begin{bmatrix}
			3$×$3,256\\
			3$×$3,256\\
			3$×$3,256\\
			\end{bmatrix}$×$1
			$}}&
	\multicolumn{1}{c|}{\multirow{6}{*}{
			$
			\begin{bmatrix}
			3$×$3,256\\
			3$×$3,256\\
			3$×$3,256\\
			\end{bmatrix}$×$1
			$}}&
	\multicolumn{1}{c|}{\multirow{6}{*}{
			$
			\begin{bmatrix}
			3$×$3,256\\
			3$×$3,256\\
			3$×$3,256\\
			\end{bmatrix}$×$2
			$}}&
	\multicolumn{1}{c|}{\multirow{6}{*}{
			$
			\begin{bmatrix}
			1$×$1,256\\
			3$×$3,256\\
			1$×$1,1024\\
			\end{bmatrix}$×$2
			$}}&
	\multicolumn{1}{c|}{\multirow{6}{*}{
			$
			\begin{bmatrix}
			1$×$1,256\\
			3$×$3,512\\
			1$×$1,512\\
			1$×$1,256\\
			3$×$3,512\\
			1$×$1,512\\
			\end{bmatrix}$×$1
			$}}
	\\
	~&~	~&~	~&~	~&~	~&~	~&~	~&~	~&~	~&~
	\\
	~&~	~&~	~&~	~&~	~&~	~&~	~&~	~&~	~&~
	\\ 
	~&~	~&~	~&~	~&~	~&~	~&~	~&~	~&~	~&~
	\\
	~&~	~&~	~&~	~&~	~&~	~&~	~&~	~&~	~&~
	\\ 
	~&~	~&~	~&~	~&~	~&~	~&~	~&~	~&~	~&~
	\\
	\hline
	\multirow{6}*{conv5x} &
	\multirow{6}*{4×4} &
	\multicolumn{1}{c|}{\multirow{6}{*}{	
			$
			\begin{bmatrix}
			3$×$3,512\\
			3$×$3,512\\
			\end{bmatrix}$×$2
			$}}&
	\multicolumn{1}{c|}{\multirow{6}{*}{
			$
			\begin{bmatrix}
			3$×$3,512\\
			3$×$3,512\\
			\end{bmatrix}$×$3
			$}}&
	
	\multicolumn{1}{c|}{\multirow{6}{*}{
			$
			\begin{bmatrix}
			1$×$1,512\\
			3$×$3,512\\
			1$×$1,2048\\
			\end{bmatrix}$×$3
			$}}&
	\multicolumn{1}{c|}{\multirow{6}{*}{
			$
			\begin{bmatrix}
			3$×$3,512\\
			3$×$3,512\\
			3$×$3,512\\
			\end{bmatrix}$×$1
			$}}&
	\multicolumn{1}{c|}{\multirow{6}{*}{
			$
			\begin{bmatrix}
			3$×$3,512\\
			3$×$3,512\\
			3$×$3,512\\
			\end{bmatrix}$×$1
			$}}&
	\multicolumn{1}{c|}{\multirow{6}{*}{
			$
			\begin{bmatrix}
			3$×$3,512\\
			3$×$3,512\\
			3$×$3,512\\
			\end{bmatrix}$×$2
			$}}&
	\multicolumn{1}{c|}{\multirow{6}{*}{
			$
			\begin{bmatrix}
			1$×$1,512\\
			3$×$3,512\\
			1$×$1,2048\\
			\end{bmatrix}$×$2
			$}}&
	\multicolumn{1}{c|}{\multirow{6}{*}{
			$
			\begin{bmatrix}
			1$×$1,512\\
			3$×$3,1024\\
			1$×$1,1024\\
			1$×$1,512\\
			3$×$3,1024\\
			1$×$1,1024\\
			\end{bmatrix}$×$1
			$}}
	\\
	~&~	~&~	~&~	~&~	~&~	~&~	~&~	~&~	~&~
	\\
	~&~	~&~	~&~	~&~	~&~	~&~	~&~	~&~	~&~
	\\ 
	~&~	~&~	~&~	~&~	~&~	~&~	~&~	~&~	~&~
	\\
	~&~	~&~	~&~	~&~	~&~	~&~	~&~	~&~	~&~
	\\ 
	~&~	~&~	~&~	~&~	~&~	~&~	~&~	~&~	~&~
	\\
	\hline
	~ & 1×1 	&\multicolumn{8}{c|}{average pool , 100-d ,fc }\\
	\hline
	\multirow{7}*{$
		\begin{matrix}
		Skip \\
		connections
		\end{matrix}
		$
	} &
	\multicolumn{1}{c|}{\multirow{7}{*}{	
			$
	    	\begin{matrix}
	    	 1 $×$1
	     	\end{matrix}
			$}}&
	\multicolumn{1}{c|}{\multirow{7}{*}{	
			$
			\begin{matrix}
			64-128,$×$1\\
			128-256,$×$1\\
			256-512,$×$1\\
			\end{matrix}
			$}}&
	\multicolumn{1}{c|}{\multirow{7}{*}{
			$
			\begin{matrix}
			64-128,$×$1\\
			128-256,$×$1\\
			256-512,$×$1\\
			\end{matrix}
			$}}&
	
	\multicolumn{1}{c|}{\multirow{7}{*}{
			$
			\begin{matrix}
			64-256,$×$1\\
			256-512,$×$1\\
			512-1024,$×$1\\
			1024-2048,$×$1\\
			\end{matrix}
			$}}&
	\multicolumn{1}{c|}{\multirow{7}{*}{
		$
		\begin{matrix}
		64-128,$×$1\\
		128-256,$×$1\\
		256-512,$×$1\\
		128-128,$×$1\\
		256-256,$×$1\\
		512-512,$×$1\\
		\end{matrix}
		$}}&
	\multicolumn{1}{c|}{\multirow{7}{*}{
		$
		\begin{matrix}
		64-128,$×$1\\
		128-256,$×$1\\
		256-512,$×$1\\
		\end{matrix}
		$}}&
	\multicolumn{1}{c|}{\multirow{7}{*}{
		$
		\begin{matrix}
		64-64,$×$ 2\\
		64-128,$×$1\\
		128-128,$×$1\\
		128-256,$×$1\\
		 256-256,$×$1\\
		  256-512,$×$1\\
		512-512,$×$1\\
		\end{matrix}
		$}}&

	\multicolumn{1}{c|}{\multirow{8}{*}{
			$
			\begin{matrix}
	      64-64, 256-64, $×$1\\
		256-128, 512-128, $×$1\\
		512-256, 1024-256, $×$1\\
		1024-512,2048-512, $×$1\\
		64-256, 128-512, $×$2 \\
		256-1024, 512-2048, $×$2 \\
			\end{matrix}
			$}}&

	\multicolumn{1}{c|}{\multirow{7}{*}{
			$
			\begin{matrix}
			64-64,$×$1\\
			128-128,$×$2\\
			256-256,$×$2\\
			512-512,$×$2\\
			1024-1024,$×$1\\
			\end{matrix}
			$}}
	\\
	~&~	~&~	~&~	~&~	~&~	~&~	~&~	~&~	~&~
	\\
	~&~	~&~	~&~	~&~	~&~	~&~	~&~	~&~	~&~
	\\
	~&~	~&~	~&~	~&~	~&~	~&~	~&~	~&~	~&~
	\\
	~&~	~&~	~&~	~&~	~&~	~&~	~&~	~&~	~&~
	\\
	~&~	~&~	~&~	~&~	~&~	~&~	~&~	~&~	~&~
	\\
	~&~	~&~	~&~	~&~	~&~	~&~	~&~	~&~	~&~
	\\
	
	\hline
	\multicolumn{2}{|c|}{FLOPS (G)} 	&
 \color{blue}$	{0.59} $& 
 \color{blue}$	{1.20} $&
 \color{blue}$	{1.34} $&
\color{orange}  $	{0.46} $&
\color{orange}  $	{0.56} $ &
\color{orange}  $	{0.92} $ &
$	0.88 $&
\color{orange}  $ {0.95}$\\
	\hline

\multicolumn{2}{|c|}{Params (M)} 	&
 \color{blue}$	{11.26} $& 
 \color{blue}$	{21.37} $&
 \color{blue}$	{23.74} $&
\color{orange}  $	{8.47} $&
\color{orange}  $	{8.56} $ &
\color{orange}  $	{17.88} $ &
$	16.27 $&
\color{orange}  $ {18.26}$\\
	\hline

\end{tabular}
\end{table*}
}

\subsection{Implementation}
We implement  ResNet* and C-ResNet models on six datasets,  i.e., MNIST, FashionMnist, CIFAR-10, CIFAR-100, CALTECH-101  and SVHN.  The same hyperparameters are used for the two types of  models.    A $32 \times 32$ crop is randomly sampled from a processed image of $40 \times 40$ by first being resized to $32 \times 32$ and then padded to $40 \times 40$.  A horizontal flip with a certain probability and normalization with a certain mean and variance are used \cite{krizhevsky2012imagenet}. We adapt BN and ReLu after each convolution \cite{ioffe2015batch}, use SGD with a weight decay of 0.0005 and a momentum of 0.9, and a learning rate starting from 0.01 and  being divided by 10 every 150 epochs. The models are trained for up to 500 epochs and a mini-batch size of 32 is used.  We run each model three times, and then take the results of the last twenty epochs for every run which generates a total of 60 results. Then we  take the mean and standard deviation of these 60 results.

\section{Experiments}  \label{Experiments}
In our experiments,  external environments (such as hyperparameter settings, optimization function etc.) on C-ResNet and ResNet* (or ResNet) are exactly the same except for the residual modules themself.  For fair comparison, we designed C-ResNets with similar layers or FLOPs  (Figs. \ref{network architecture-1} and \ref{network architecture-2})  to compare performance of ResNets. We use ResNet* as the baseline in our experiments not only because  its classification accuracy on datasets CIFAR-10 and CIFAR-100 is much higher than that of ResNet (\cite{he2016identity}) (Tab. \ref{ResNet and ResNet*}), but also because the BasicBlock and Bottleneck modules in  ResNet* are exactly the same as in ResNet.

\setlength{\tabcolsep}{3mm}{ 
\begin{table}[!htbp]\small 
\caption{The error rate comparison between ResNets and fine-tuned ResNets. \\ The results with \# are obtained from \cite{he2016identity}, and those without \# are produced by our experiments.}
\label{ResNet and ResNet*}
\centering

\begin{tabular}{ccc} 
\toprule 
& CIFAR-10 & CIFAR-100 \\
\hline
Resnet100$^\#$ & 6.37 & - \\
Resnet164$^\#$& 5.46  & 24.33  \\
Resnet1001$^\#$ & 4.92  & 22.71  \\

\hline
Fine-tuned Resnet18 &  4.52  &  21.42 \\
Fine-tuned Resnet34  & 4.25  &  20.23 \\
Fine-tuned Resnet50 & 3.96 & 19.45 \\

\bottomrule 

\end{tabular}
\end{table}
}

\subsection{The C-ResNets designed and evaluated} \label{The C-ResNets designed and evaluated}

\subsubsection{C-ResNet15}
C-ResNet15 (Fig. \ref{C-ResNet15-A1})  is constructed to compare the performance with fine-tuned ResNet18 (i.e., ResNet18*, see Fig. \ref{ResNet18}). ResNet18*  has 18 conv layers where the size of the conv kernel of each layer is $3 \times 3$ and 8 no-cross skip connections (or jumpers) including 5 solid jumpers and 3 dashed jumpers. 

\begin{figure*}[htb]
 \subfigure[Flops]{
  \label{flops comparison}      
  \begin{minipage}[b]{0.5\textwidth}
   \centering
   \includegraphics[width=3in]{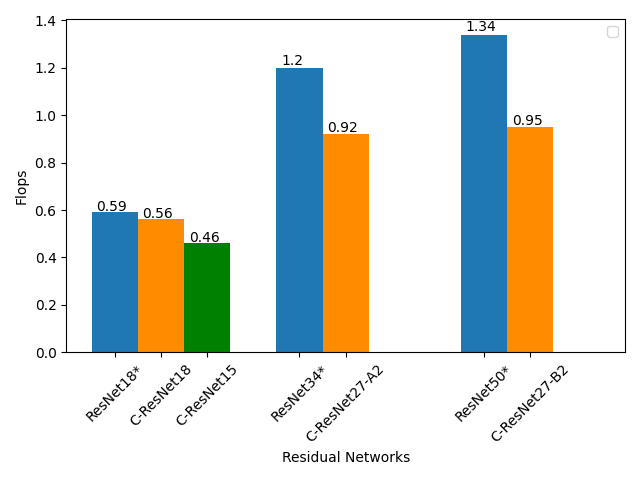}
  \end{minipage}}%
\subfigure[The number of parameters]{
  \label{parameters comparison}       
  \begin{minipage}[b]{0.5\textwidth}
   \centering
   \includegraphics[width=3in]{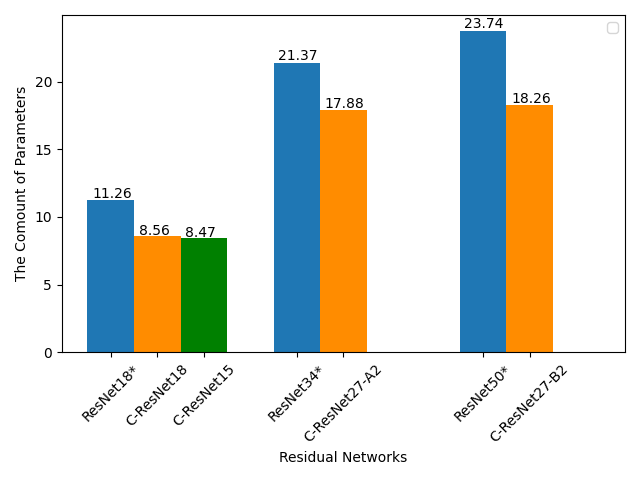}
  \end{minipage}}%
 \caption{The costs comparison between C-ResNets and ResNets* }
 \label{The costs comparison between C-ResNets and ResNets*} 
\end{figure*}

C-ResNet15  has 14 conv layers where each conv kernel is $3 \times 3$ and 8 cross  jumpers including 2 solid jumpers and 6 dashed jumpers. Compared with ResNet18*, C-ResNet15 achieves a reduction in computational costs (i.e., FLOPs) and the amount of parameters by 22.03\% and 24.78\% (Fig. \ref{The costs comparison between C-ResNets and ResNets*}), respectively,  and has lower error rates on five datasets (Tab. \ref{accurate comparison table}), i.e.,  MNIST, FashionMnist, CIFAR-10, CALTECH-101 and SVHN. In particular, the error rate of C-ResNet15 on the dataset CALTECH-101 is 7.96\% lower than ResNet18*, and its variance is also smaller indicating that it is more stable.

\setlength{\tabcolsep}{1.2mm}{ 
\begin{table*}[!htbp]\small 
\caption{Test error rate comparison between C-ResNets and ResNets*.(\%)}
\label{accurate comparison table}
\centering

\begin{tabular}{cccccccc} 
\toprule 
\diagbox{model}{test error (\%)}{datasets}& MNIST & FashionMnist  &  CIFAR-10  & CIFAR-100 & CALTECH-101  & SVHN \\
\hline
ResNet18* & ${0.44\pm0.02}$ &    ${6.85\pm0.07}$    & ${4.52\pm0.13}$ & \color{orange} ${21.42\pm0.19}$ & ${16.96\pm0.52}$ & ${3.26\pm0.08}$ \\
C-ResNet15 & \color{orange} ${0.41\pm0.05}$ &  \color{orange} ${6.66\pm0.08}$  &\color{orange} ${4.28\pm0.13}$ &   ${21.49\pm0.16}$ & \color{orange} ${15.61\pm0.5}$ &\color{orange}  ${3.24\pm0.07}$ \\ 

\hline
ResNet18* & ${0.44\pm0.02}$ &    ${6.85\pm0.07}$    & ${4.52\pm0.13}$ & ${21.42\pm0.19}$ & ${16.96\pm0.52}$ & ${3.26\pm0.08}$ \\
C-ResNet18 &\color{brown} ${0.40\pm0.02}$ & \color{brown}  ${6.68\pm0.10}$  &\color{brown}  ${4.36\pm0.06}$ & \color{brown} ${21.1\pm0.16}$ & \color{brown}  ${16.52\pm0.62}$ & \color{brown} ${3.22\pm0.06}$ \\ 

\hline
ResNet34* & ${0.49\pm0.05}$   &   ${7.17\pm0.13}$ & ${4.25\pm0.05}$ & \color{SeaGreen3} ${20.23\pm0.09}$ &\color{SeaGreen3} ${15.95\pm0.57}$ & ${3.14\pm0.14}$ \\
C-ResNet27-A2 &  \color{SeaGreen3}  ${0.44\pm0.02}$    & \color{SeaGreen3}  ${7.01\pm0.11}$   & \color{SeaGreen3} ${4.23\pm0.1}$ &${21\pm0.12}$ &  ${16\pm0.46}$ & \color{SeaGreen3} ${3.13\pm0.04}$ \\ 
\hline

ResNet50*  &${0.45\pm0.03}$    &   \color{blue}  ${6.39 \pm0.09}$  & \color{blue} ${3.96\pm0.1}$ & \color{blue} ${19.45\pm0.12}$ & ${16.79\pm0.45}$ & ${3.22\pm0.09}$ \\
C-ResNet27-B2  &  \color{blue}  ${0.41\pm0.03}$     &    ${6.6 \pm0.06}$    &   ${4.22\pm0.15}$        &      ${20.1\pm0.20} $   & $\color{blue}{16.34\pm0.47}$ & \color{blue} ${3.21\pm0.05}$ \\ 
\bottomrule 

\end{tabular}
\end{table*}  
}

Furthermore, C-ResNet15  has the same jumpers number as ResNet18*, but it has higher proportion of dashed jumpers to jumpers and  fewer $3\times3$ conv layers than ResNet18*. This means that we may cross the jumpers and increase the proportion of dashed jumpers in residual network to improve its classification accuracy while reducing its number of layers. Compared with the increased cost of the dashed jumpers, the reduced conv layers drop more cost. Therefore, we have reduced the total resource cost including the amount of calculation and parameters.

\subsubsection{C-ResNet18}
C-ResNet18 (Fig. \ref{C-ResNet18-A})  is also constructed to compare the performance with ResNet18*. C-ResNet18  has 17 conv layers and 10 cross  jumpers including 7 solid jumpers and 3 dashed jumpers. Compared with ResNet18*, it drops in computational costs and the amount of parameters by 5.08\% and 23.98\% (Fig. \ref{The costs comparison between C-ResNets and ResNets*}), respectively,  and has lower error rates on all six datasets  (Tab. \ref{accurate comparison table}). We can see that C-ResNet18 greatly reduces the amount of parameters so that the utilization rate of the GPU memory can be dramatically cut down.

C-ResNet18  has fewer conv layers and more jumpers (only 2 more solid jumpers) than ResNet18*, implying that we can cross jumpers to make jumpers densification increasing information retention and delivery in residual network while reducing the conv layer. Therefore, jumpers densification in residual network may only increasing solid jumpers (addition operation is implemented and the increased resource cost is almost negligible) can improve the classification performance while decreasing the amount of calculation and parameters.

\subsubsection{C-ResNet27-A2}
C-ResNet27-A2 (Fig. \ref{C-ResNet27-A2})  is constructed through stacking more WBR modules (a 2-2-2-2 structure) in order to compare the performance with fine-tuned ResNet34 (i.e., ResNet34*).  ResNet34*  (Tab. \ref{The architectures for our C-ResNet and ResNet*}) has 34 conv layers and 16 no-cross skip connections (or jumpers) including 13 solid jumpers and 3 dashed jumpers.

C-ResNet27-A2 has 26 conv layer and 16 cross jumpers including 8 solid jumpers and 8 dashed jumpers, which separately realizes  23.33\%  and  16.33\%  reduction in computational costs and  amount of parameters  compared with  ResNet34* (Fig. \ref{The costs comparison between C-ResNets and ResNets*}). Meanwhile, C-ResNet27-A2 offers lower error rates than ResNet34* on datasets MNIST, FashionMnist, CIFAR-10 and SVHN and also shows stronger stability on datasets MNIST, FashionMnist, CALTECH-101 and SVHN due to smaller variances  (Tab. \ref{accurate comparison table}).

We find that performance difference between C-ResNet27-A2 and ResNet34* on different datasets 
are fairly close and each model shows own advantages and disadvantages, but the difference is insignificant. However, the huge reduction in computational costs and amount of parameters in C-ResNet27-A2 greatly abates the utilization rate of GPUs and GPU memory  such that C-ResNet27-A2  shows a greater strength than ResNet34*. 

C-ResNet27-A2  has the same number of jumpers as  ResNet34*, but  fewer conv layers and more dashed jumpers, which further strengthens this conclusion that we can cross jumpers and convert solid jumpers to dashed jumpers such that  the number of conv layers can be reduced without reducing the feature diversity.

\subsubsection{C-ResNet27-B2}
C-ResNet27-B2 based on a variant of six-layer Cross-Bottleneck module (Fig. \ref{Cross-Bottleneck-B}) where the second solid jumper becomes dashed jumper, is constructed to compare with ResNet50*. ResNet50* has 50 conv layers and 16 no-cross jumpers including 13 solid jumpers and 3 dashed jumpers.

C-ResNet27-B2 has 26 conv layers  which include $3 \times 3$  and  $1 \times 1$ conv layers,  and 12 cross jumpers including 4 solid jumpers and 8 dashed jumpers. Compared with ResNet50*, C-ResNet27-B2 has higher accuracy on three datasets and lower accuracy on the other three datasets. Therefore, C-ResNet27-B2 is competitive in classification accuracy. However, due to its huge reduction in the amount of calculations and parameters (respectively by 29.10\% and 23.08\%), it offers obvious advantages over ResNet50*. 

C-ResNet27-B2 has fewer conv layers, fewer total jumpers and more dashed jumpers than ResNet50*,  which further substantiates our conjecture on the function of crossing  jumpers and using dashed jumpers instead of solid jumpers. Through the two methods,  we can increase the density of jumpers and maintain the feature diversity such that  we can reduce the number of conv layers and computational complexity without decreasing classification performance.

\subsubsection{Conclusion}
Compared with fine-tuned  ResNet18, ResNet34 and ResNet50 on datasets MNIST, FashionMnist, CIFAR-10, CIFAR-100, CALTECH-101 and SVHN,  our corresponding cross-residual networks C-ResNet15 (or C-ResNet18), C-ResNet27-A2 and  C-ResNet27-B2  offer superior performance in terms of computational costs (Fig. \ref{flops comparison}), amount of parameters (Fig. \ref{parameters comparison}) and classification accuracy  (Tab. \ref{accurate comparison table}). The above experiment results infer that densifying the jumpers (through crossing jumpers) and changing the role of jumpers (that is, the solid jumpers are changed into dashed jumpers, not to match the size of the two ends of jumpers, but to increase the interaction of information) are two important strategies for residual networks to lower the resource cost while maintaining or even improving the performance.

\subsection{The deployment strategy for  jumpers} \label{The deployment strategy for  jumpers}

When designing cross jumpers for our framework, we introduce two types of operations, that is,
one (solid jumpers) is identity mapping and the other (dashed jumpers) is a \(1\times 1\) conv.
The  \(1\times 1\) conv operation not only serves to match input-output dimensions, but helps to enhance feature diversity by increasing the feature transformations involved.
In other words, even if the dimensions on both ends match exactly, it may still be beneficial to introduce such a dashed jumper. 
So is the number of dashed jumpers as many as possible? Whether more numbers can better reflect the diversity of features.
In this section, we further discuss the impact of  the deployment strategy of jumpers on the classification performance of C-ResNets on datasets CIFAR-10, CIFAR-100 and CALTECH-101.

\setlength{\tabcolsep}{3mm}{ 
	\begin{table*}[!htbp]\small 
		\caption{The test error comparison for C-ResNets having different dashed jumper schemes. (\%)}
		\label{skip connection comparison}
		\centering

	\begin{tabular}{ccccc}   
	\toprule  
     & params & CIFAR-10 & CIFAR-100 & CALTECH-101 \\
      \hline
  C-ResNet27-A & \color{orange} 17.53M & ${4.31\pm0.13}$ & ${21.01\pm0.29}$ & ${25.77\pm1.08}$  \\
  C-ResNet27-A1 & 17.87M & ${4.29\pm0.12}$ & ${21.35\pm0.23}$ & ${26.81\pm0.92}$ \\
C-ResNet27-A2 & 17.88M & \color{orange}${4.23\pm0.1}$   & \color{orange}${21\pm0.12}$  & \color{orange}${25.52\pm0.89}$    \\
      \hline
 C-ResNet27-B & 16.872M   &   ${4.14\pm0.06}$      &     ${20.04\pm0.18} $     & ${16.5\pm0.73}$ \\
  C-ResNet27-B1 & \color{blue} {16.868M}   &   \color{blue}   ${4.11\pm0.06}$         &   \color{blue}  ${19.96\pm0.13} $     & ${16.21\pm0.45}$ \\
   C-ResNet27-B2 & 18.26M   &    ${4.22\pm0.15}$        &      ${20.1\pm0.20} $       & ${16.34\pm0.47}$ \\
    C-ResNet27-B3 &  19.66M   &   ${4.26\pm0.22}$       &     ${20.09\pm0.24} $      & \color{blue}  ${15.93\pm0.74}$ \\
\hline
C-ResNet27-C1 &  \color{SeaGreen3} 16.27M &   ${4.53\pm0.15}$      &     ${21.58\pm0.32} $     & ${16.96\pm0.49}$  \\
 C-ResNet27-B2 & 18.26M   &     \color{SeaGreen3} ${4.22\pm0.15}$        &     \color{SeaGreen3}  ${20.1\pm0.20} $      &    \color{SeaGreen3} ${16.34\pm0.47}$ \\
\bottomrule 
		\end{tabular}
	\end{table*}
}

\subsubsection{ C-ResNet27-A series}

Leveraging Cross-Block (Fig. \ref{Cross-Block-A}) and its variants, we construct  three C-ResNet27 (2,2,2,2)-networks  A, A1 and A2 (Fig. \ref{network architecture-1}), which have the same network structures except that the ratio of solid jumpers to dashed jumpers is different, and have three, six and eight dashed jumpers, respectively.

   \begin{figure*}[htb]
    \subfigure[C-ResNet15-A1 (also called C-ResNet15)]{
  \label{C-ResNet15-A1}       
  \begin{minipage}[b]{0.22\textwidth}
   \centering
   \includegraphics[width=1.31in]{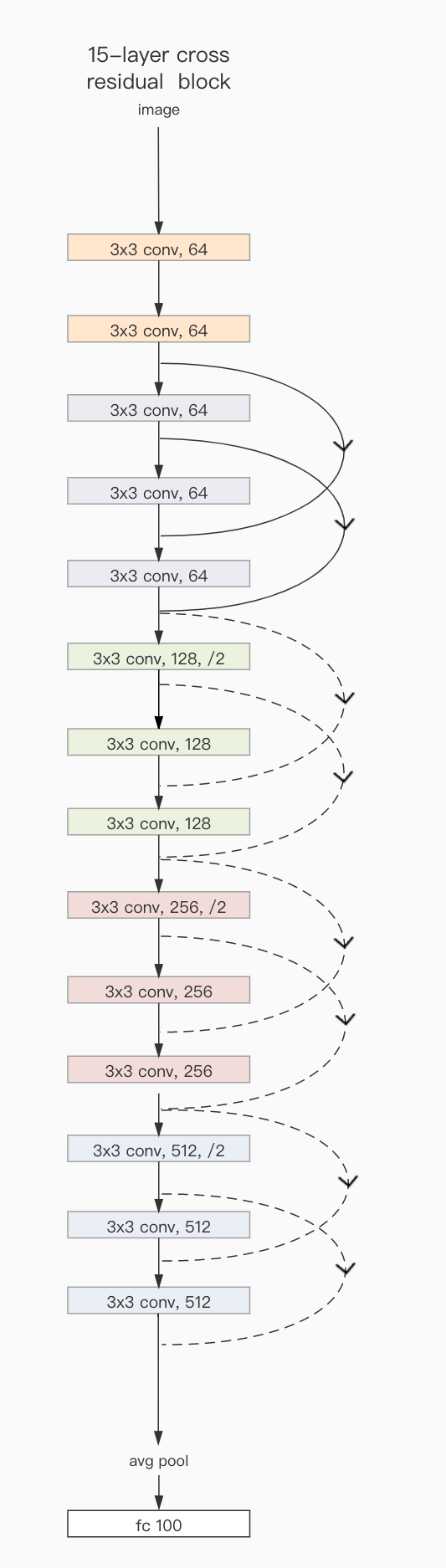}      
    \end{minipage}}%
\subfigure[C-ResNet27-B]{
  \label{C-ResNet27-B}       
  \begin{minipage}[b]{0.2\textwidth}
   \centering
   \includegraphics[width=1.19in]{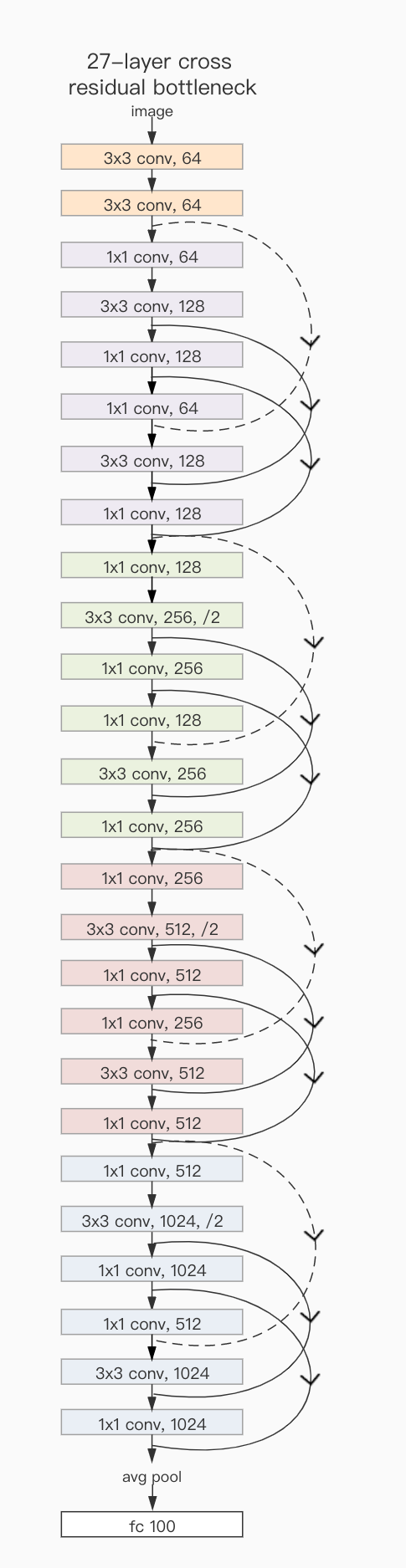}  
  \end{minipage}}%
  \subfigure[C-ResNet27-B1]{
  \label{C-ResNet27-B1}       
  \begin{minipage}[b]{0.2\textwidth}
   \centering
   \includegraphics[width=1.18in]{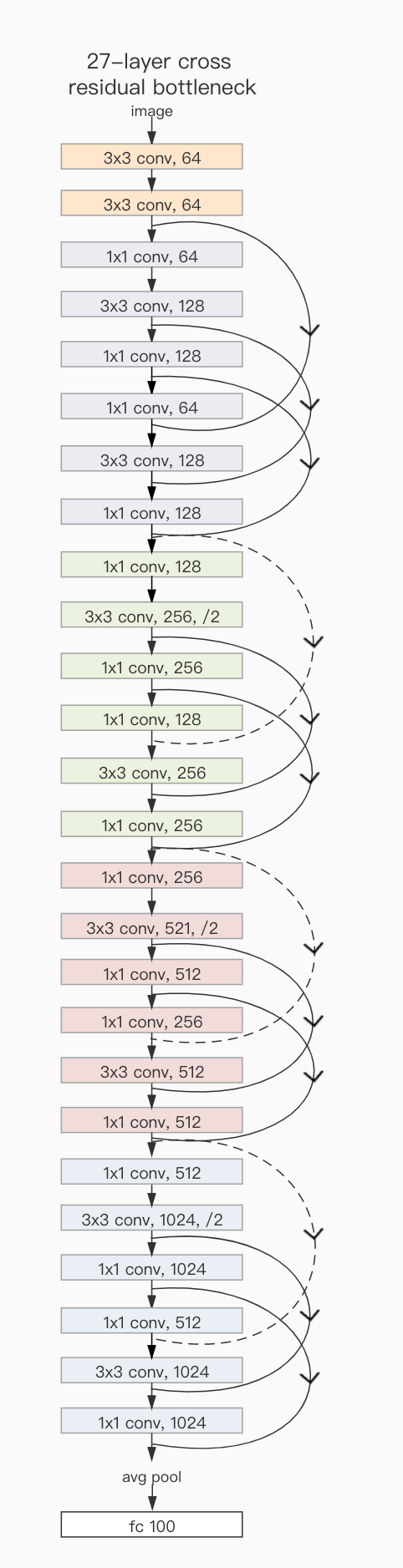}  
  \end{minipage}}%
  \subfigure[C-ResNet27-B2]{
  \label{C-ResNet27-B2}       
  \begin{minipage}[b]{0.2\textwidth}
   \centering
   \includegraphics[width=1.18in]{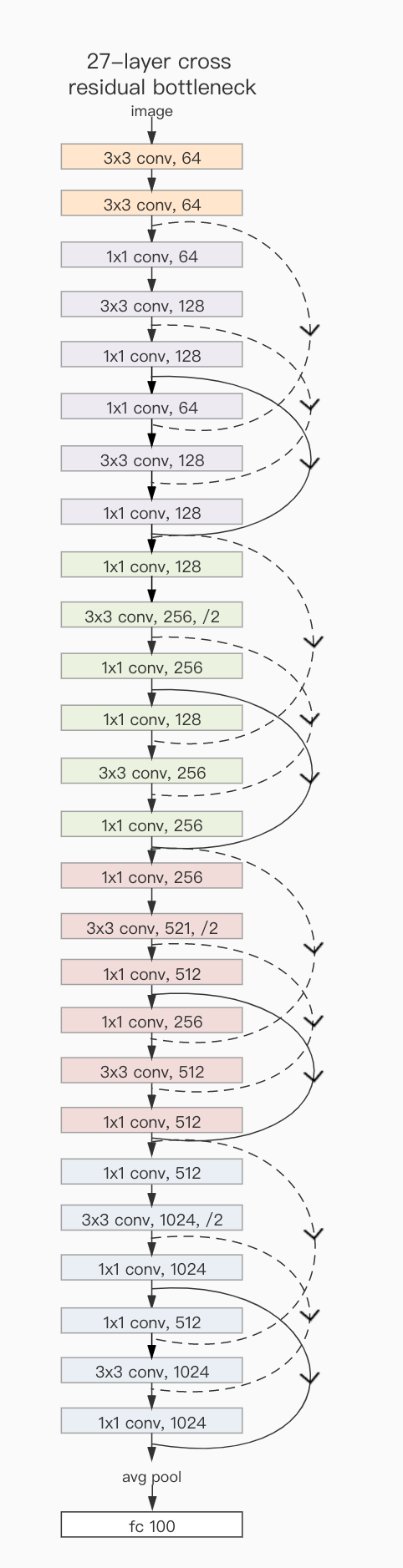}  
  \end{minipage}}%
  \subfigure[C-ResNet27-B3]{
  \label{C-ResNet27-B3}       
  \begin{minipage}[b]{0.2\textwidth}
   \centering
   \includegraphics[width=1.18in]{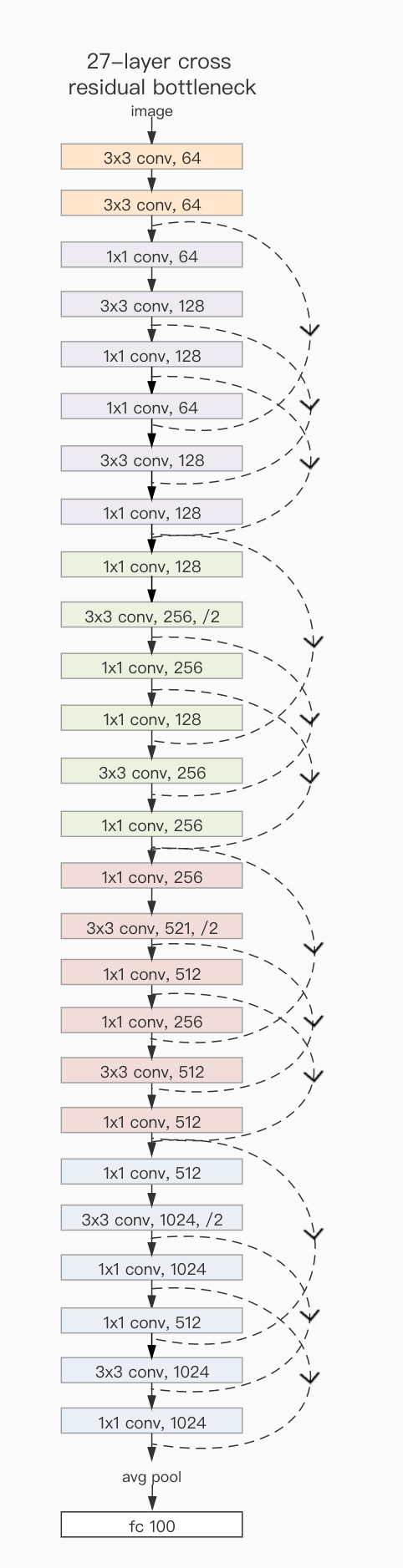}  
  \end{minipage}}%
 \caption{C-ResNets samples} 
 \label{network architecture-2} 
\end{figure*}

Compared with A, A1 has more dashed jumpers, but its classification accuracy on datasets CIFAR-100 and CALTECH-101 are reduced (Tab. \ref{skip connection comparison}). However, comparing with A and A1, A2 has more dashed jumpers and higher classification accuracy (Tab. \ref{skip connection comparison}). Therefore, it should also be pointed out that arbitrarily incorporating dashed jumpers will not always bring performance gain. We observe that the number and structure of dashed  jumpers are both important in C-ResNets.

\subsubsection{C-ResNet27-B series}
Based on Cross-Bottleneck (Fig. \ref{Cross-Bottleneck-B}) and its variants,  four  C-ResNet27  (1,1,1,1)-networks (Fig. \ref{network architecture-2}) B, B1,B2 and B3 are constructed which  have the same network structure except for different ratio of solid jumpers to dashed jumpers, and have four, three, eight and twelve dashed jumpers, respectively. 

From middle parts of Tab. \ref{skip connection comparison}, we  can also see that  B3 (twelve dashed jumpers), which has the largest number of dashed jumpers and the highest classification accuracy on the dataset CALTECH-101, but the classification accuracy on the datasets CIFAR-10 and CIFAR-100 is not the highest.

Therefore, C-ResNets with more dashed jumpers (e.g., B3) may not have higher classification accuracy on some datasets, which further verified the importance for the number and structure of dashed jumpers in C-ResNets.

\begin{figure*}[htb]
 \subfigure[~]{
  \label{b2-c1-Mnist}      
  \begin{minipage}[b]{0.5\textwidth}
   \centering
   \includegraphics[width=0.7\textwidth]{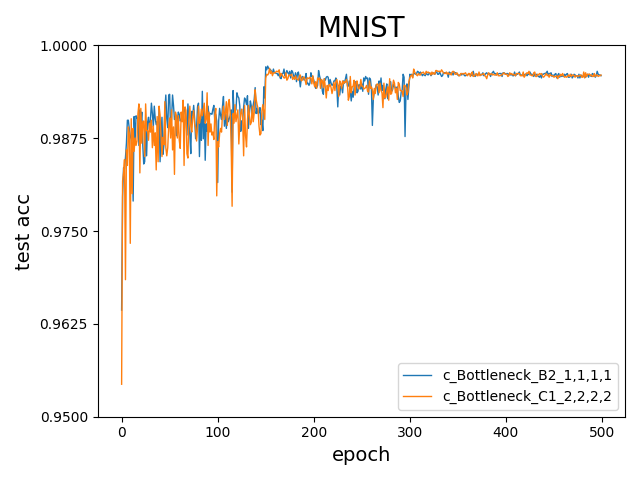}
  \end{minipage}}%
\subfigure[~]{
  \label{b2-c1-FashionMnist}       
  \begin{minipage}[b]{0.5\textwidth}
   \centering
   \includegraphics[width=0.7\textwidth]{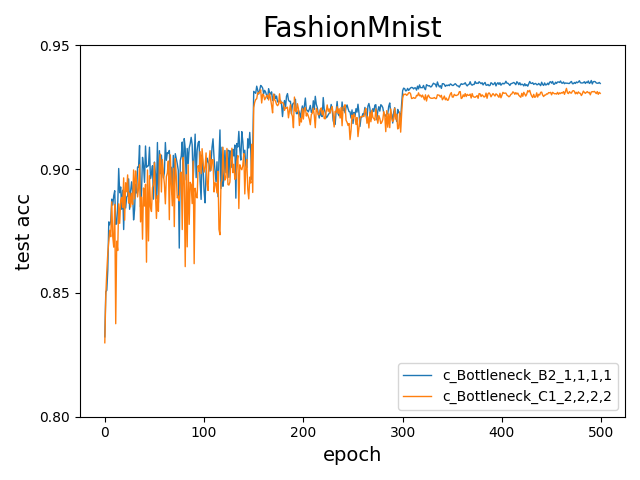}
  \end{minipage}}%
  
  \subfigure[~]{
  \label{b2-c1-Cifar10}       
  \begin{minipage}[b]{0.5\textwidth}
   \centering
   \includegraphics[width=0.7\textwidth]{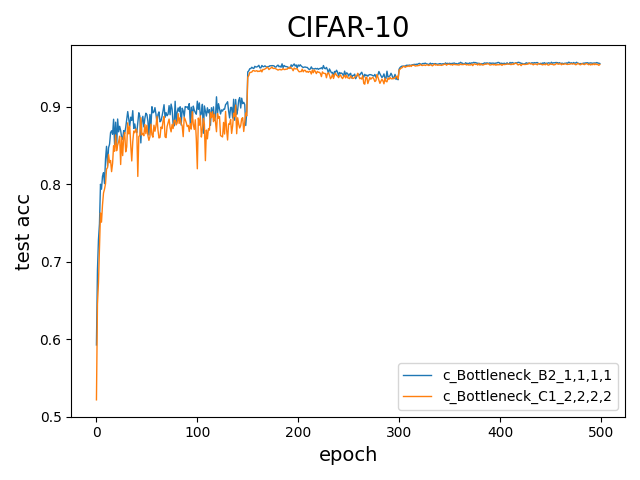}
  \end{minipage}}%
    \subfigure[~]{
  \label{b2-c1-Cifar100}       
  \begin{minipage}[b]{0.5\textwidth}
   \centering
   \includegraphics[width=0.7\textwidth]{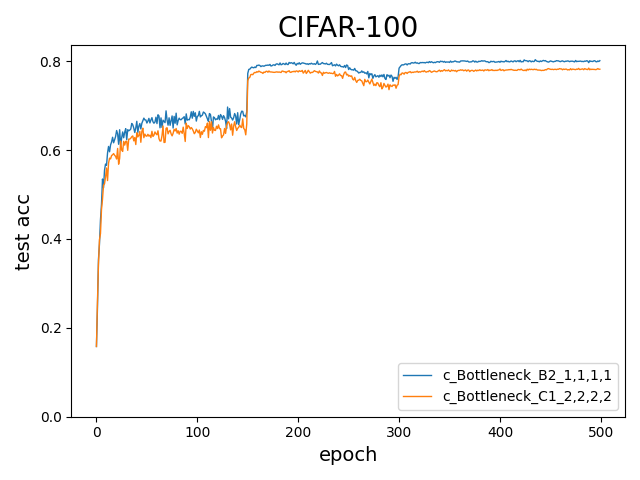}
  \end{minipage}}%
  
    \subfigure[~]{
  \label{b2-c1-Caltech101}       
  \begin{minipage}[b]{0.5\textwidth}
   \centering
   \includegraphics[width=0.7\textwidth]{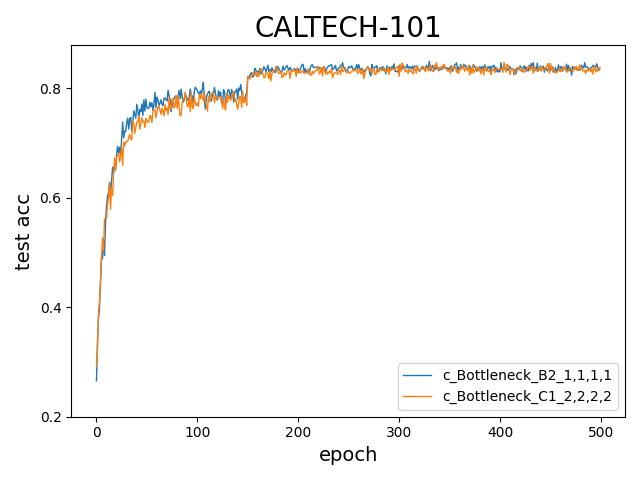}
  \end{minipage}}%
    \subfigure[~]{
  \label{b2-c1-SVHN}       
  \begin{minipage}[b]{0.5\textwidth}
   \centering
   \includegraphics[width=0.7\textwidth]{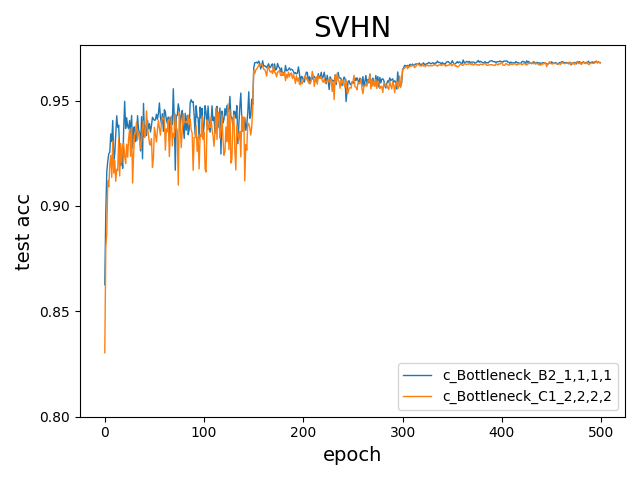}
  \end{minipage}}%
 \caption{The test  accuracy comparison between C-ResNet-B2 and C-ResNet-C1. }
 \label{b2-c1-test-acc} 
\end{figure*}

\subsection{The trade-off between channel and dashed jumpers}
From the design of Bottleneck module (Fig. \ref{Original-Bottleneck-E}) in ResNet \cite{he2016deep}, it may be tempting to derive a similar design scheme for three-layer Cross-Bottleneck module  (Fig. \ref{Cross-Bottleneck-C}) in C-ResNet. However, when we build C-ResNet based on this three-layer modules, such a scheme can easily cause all jumpers to be dashed jumpers due to the inconsistency of channels  (e.g., C-ResNet27-C1, shown in github repository).Such C-ResNet may not perform optimally.  Hence,  a six-layer Cross-Bottleneck module (Fig. \ref{Cross-Bottleneck-B}) is designed  to maintain a certain ratio between the solid jumpers and the dashed jumpers.

C-ResNet27-B2  and C-ResNet27-C1 are constructed separately based on the six-layer Cross-Bottleneck and the three-layer Cross-Bottleneck, which have the same network structure and the number of conv layers and conv kernels.  Compared with C-ResNet27-C1, C-ResNet27-B2 has fewer dashed jumpers,  but has more channels in the $3 \times3$ conv layers. Therefore, it can be regarded as C-ResNet27-B2 makes a trade-off between channel and jumper.

It can be seen that C-ResNet27-B2 has higher classification accuracy except for on MNIST (Fig. \ref{b2-c1-test-acc}), but higher parameter complexity and computational cost than C-ResNet27-C1  (Tab. \ref{The architectures for our C-ResNet and ResNet*}). Although the cost is increased, the classification performance is also improved. Therefore,  it is feasible to increase channels to help enhance feature diversity while reducing the dashed jumpers in C-ResNets.

\subsection{Object Detection on MS COCO}
We use the official code from Pytorch \footnote{The code is available at \url{https://github.com/pytorch/vision/tree/main/torchvision/models/detection} and \url{https://github.com/pytorch/vision/tree/main/references/detection}} to achieve an end-to-end training of ResNet without loading pre-trained parameters. The execution settings and peripheral environment of ResNet and C-ResNet is exactly the same except for the stacked modules.  We evaluate our models with metric mAP @IoU = 0.5 and mAP @ IoU =[.5, .95]. Tab. \ref{Object Detection} show the object detection results on MS COCO dataset \cite{lin2014microsoft}. Compared with ResNet18 and ResNet50, the performance of C-ResNets has only a small gap; compared with ResNet34, the performance of C-ResNet27-A2 slightly exceeds it on mAP @IoU = 0.5. Therefore, C-ResNet has also good generalization performance as ResNet on object detection task.

\begin{table}[]
\caption{Object detection on MS COCO}
\label{Object Detection}
	\centering
\begin{tabular}{ccc}
\toprule 
   & @.5            & @[.5,.95]      \\
\hline
ResNet50      & 0.435          & 0.259          \\
C-ResNet27-B2& 0.423          & 0.245          \\
\hline
ResNet34      & 0.447          &  \color{blue}{0.273} \\
C-ResNet27-A2  &  \color{blue}{0.448} & 0.272          \\
 \hline
ResNet18      & 0.422          & 0.249          \\
C-ResNet18    & 0.421          & 0.248          \\
C-ResNet15  & 0.373          & 0.207          \\ 
\bottomrule 
\end{tabular}
\end{table}

\subsection{Discussion}

\cite{bello2021revisiting} and \cite{liu2022convnet} have experimentally verified that after adding fine training tricks and several key components from the current SOTA, the performance of ResNet can be greatly improved. C-ResNet can be regarded as the similar backbone as ResNet. Therefore,  after adding fine training tricks and several key components, C-ResNet can also be a competitive model compared with SOTA method, such as Vision Transformer \cite{dosovitskiy2020image}, Swin Transformer \cite{he2022masked}, ConvNeXt \cite{liu2022convnet} and MAE\cite{he2022masked}.

\section{Conclusion and future work} 
A cross-residual learning framework is proposed  for conv neural networks built by stacking blocks containing multiple cross jumpers.
Jumper densification facilitates information forwarding from earlier layers to subsequent ones, which can lead to better feature diversity, much lower computational costs and much lower memory consumption.  
Jumper density, deployment scheme of dashed jumpers and the channel counts are three important factors impacting C-ResNets performance.
We argue that three strategies, i.e.,  increasing jumper density,  maintaining a suitable ratio as well as structure for the solid and dashed jumpers and increasing the number of channels,  can strengthen the interaction between information, maintain the diversity of features and improve the performance in most residual networks. 

We have designed several basic residual modules and constructed several cross residual networks (C-ResNets) in this paper. These basic residual modules can be directly applied in the construction of various convolutional neural networks and these C-ResNets can also be utilized to accomplish various tasks of computer vision due to their low consumption and high performance. Furthermore, the three strategies that have a greater impact on the performance of residual networks can be applied to construct various neural networks containing residual ideas in different fields.

The more refined design and stacking of the cross residual module can further reduce resource cost of cross residual networks and improve its performance. In addition, derived from the variant ResNeSt of ResNet, we believe that  widening the width of modules can gain further performance enhancements of C-ResNets. Furthermore, the cross residual modules may also be combined with other ideas (such as multiscale, adaptive dynamics learning, wider blocks, attention mechanism, training methods and scaling strategies \,  etc.) to play an important role in specific fields. Therefore, fine design of  the cross residual module,  widening the width of modules (concatenating the feature maps) and adding attention mechanism are our future work.

\section*{Acknowledgments}
This work was partially supported by Guangdong Provincial Basic and Applied Basic Research Fund Regional Joint Fund - Regional Cultivation Project, Guangdong Basic and Applied Basic Research Fund Regional Joint Fund Project (Key Project) (2020B1515120089), Foshan Higher Education High-level Talent Project.

\bibliography{Deep-learning}

\end{document}